\pdfoutput=1
\documentclass[11pt]{article}

\usepackage{EMNLP2022}

\usepackage{times}
\usepackage{latexsym}

\usepackage[T1]{fontenc}
\usepackage[utf8]{inputenc}

\usepackage{microtype}

\usepackage{inconsolata}

\usepackage{amsmath}
\usepackage{amssymb}
\usepackage{graphicx}
\usepackage{multirow}
\usepackage{float}

\title{Mixed-Effects Transformers for Hierarchical Adaptation}

\author{
  Julia White \\
  Electrical Engineering \\
  Stanford University\\
  \small{\texttt{jiwhite@stanford.edu}} 
  \\\And
  Noah D. Goodman \\
  Computer Science, Psychology \\
  Stanford University \\
  \small{\texttt{ndg@stanford.edu}} \\
  \\\And
  Robert D. Hawkins \\
  Psychology \\
  Princeton University \\
  \small{\texttt{rdhawkins@princeton.edu}}
}

\begin{document}
\maketitle
\begin{abstract}
Language differs dramatically from context to context. To some degree, large language models like GPT-3 account for such variation by conditioning on strings of initial input text, or \emph{prompts}.  However, prompting can be ineffective when contexts are sparse, out-of-sample, or extra-textual. In this paper, we introduce the \emph{mixed-effects transformer} (MET), a novel approach for learning hierarchically-structured prefixes--- lightweight modules prepended to an input sequence--- to account for structured variation in language use. Specifically, we show how the popular class of mixed-effects regression models may be extended to transformer-based architectures using a regularized prefix-tuning procedure with dropout. We evaluate this approach on several domain-adaptation benchmarks, finding that it learns contextual variation from minimal data while generalizing well to unseen contexts.
\end{abstract}

\section{Introduction}

While certain aspects of language use are nearly universal -- such as basic grammatical acceptability \cite{warstadt2019neural,linzen2021syntactic} or simple lexical judgements \cite{wang2019superglue} -- these often seem to be the exception that proves the rule. 
Contextual variation is ubiquitous in language, where predictions may differ as a function of speaker identity \cite{blodgett2016demographic,yang2017overcoming,Ostapenko2022}, location \cite{hofmann2022geographic}, time \cite{lazaridou2021mind,sawhney2020time,schlechtweg-etal-2019-wind,rottger2021temporal}, or usage domain \cite{dai-etal-2020-cost,nguyen2020bertweet,lee2020biobert}.
Although such variation has long been recognized in psycholinguistics \cite{Clark1998-CLACL} and sociolinguistics \cite{nardy2013acquisition,eckert2012three}, the dominant approach in modern NLP has been to train monolithic models  \cite{flek2020returning,hovy2015demographic} and fine-tune for individual domains if necessary \cite[e.g.][]{daume2006domain}.

\begin{figure}[t!]
\begin{center}
\includegraphics[width=7.5cm]{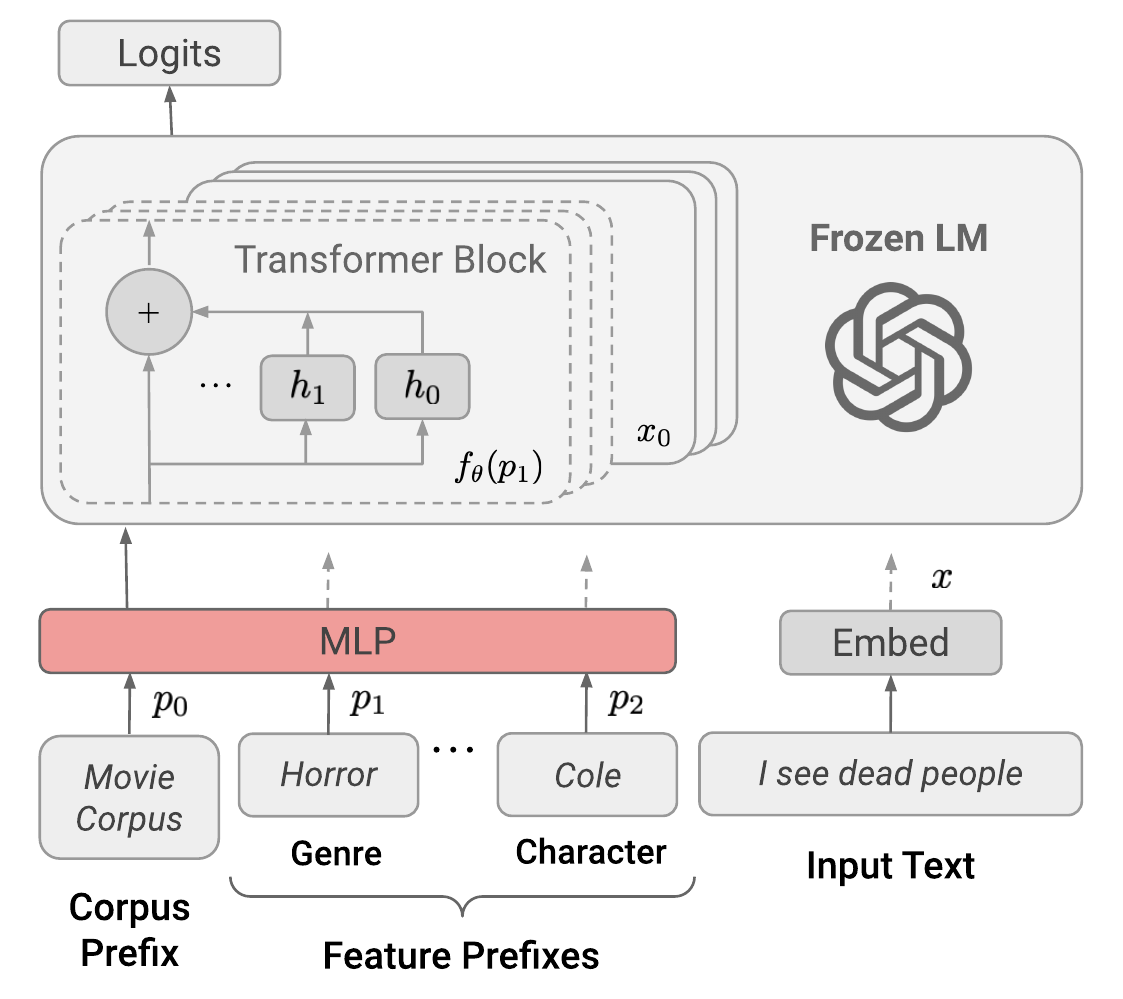}
\end{center}
\caption{In the mixed-effects transformer (MET), parameters of a pretrained transformer are frozen (solid border) while prefixes are adapted to different contextual features (dashed border).}
\label{arch}
\end{figure}

Recent large language models (LLMs) like \mbox{GPT-3}  \cite{Brown2020,bommasani2021opportunities} have begun to provide a more systematic approach for handling context-specific variance.
By adding relevant contextual information to the text input (i.e. prompting), these models have been able to account for known demographic information such as the speaker's age, gender, or country of origin \cite{Ostapenko2022}.
However, it is less clear how to use prompting when context is extra-textual, contains multiple features, or lies outside the training distribution.
For example, LLMs trained prior to the COVID-19 pandemic failed catastrophically on the torrent of new tweets and medical papers \cite{feldman2021analyzing,zeng-etal-2020-meddialog,luu2021time}.

In these cases, some degree of online adaptation is required.
One particularly promising adaptation technique is \emph{prefix-tuning}, where a lightweight module is prepended to the input and fine-tuned to modulate a downstream network that has been frozen \cite{Li2021}. 
To date, however, this technique has only been used to fine-tune prefixes for distinct downstream tasks \cite[see also][]{hambardzumyan2021warp,zhou2021learning,lester2021power}.
In this paper, we suggest that the prefix-tuning approach is particularly well-suited for \emph{hierarchical} adaptation in language modeling.
Specifically, we show how a form of dropout may be used to implement \emph{random effects}, yielding a \emph{mixed-effects transformer} (MET; \autoref{arch}).
This approach allows the model to learn strong domain-specific predictions for frequently occurring prefixes while abstracting away generalizable inductive biases for sparser or unseen contexts. Our code is available at  \url{https://github.com/juliaiwhite/mixed-effects-transformers}.

% Domain structure has become increasingly prevalent as language modeling datasets have grown to incorporate language data from many distinct sources

\section{Approach}

We begin by reviewing  mixed-effects models in a classic hierarchical regression setting before extending it to explicitly model contextual variation with modern language models.

\begin{figure}[t!]
\begin{center}
\includegraphics[width=6cm]{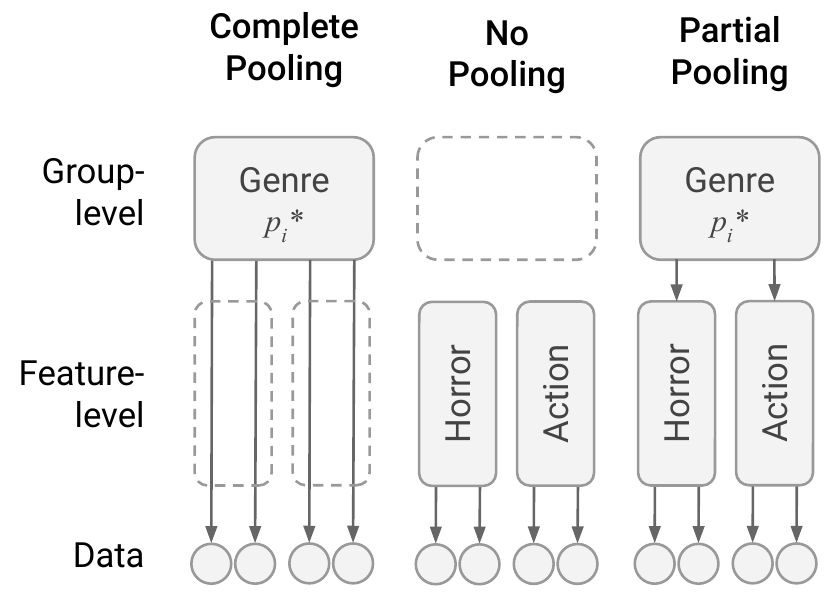}
\end{center}
\caption{Complete pooling approaches learn a single model representing the central tendency across all domains while no pooling learns separate models for each domain. Mixed-effects models combine the two.}
\label{pooling}
\end{figure}

\paragraph{Mixed-effects regression.}
Mixed-effects models, also known as multi-level models or partial pooling models, may be understood as a way of interpolating between two extremes which are each prevalent in machine learning \cite{gelman2006data,Baltagi2021,hawkins2021partners}, as illustrated in \autoref{pooling}. On one hand, \emph{complete-pooling} approaches learn a single monolithic model across multiple domains, thus generalizing well to out-of-distribution data. \emph{No-pooling} approaches, on the other hand, learn separate models for each domain, enabling stronger in-distribution predictions. 

Mixed-effects models offer a balance between these approaches by combining \emph{fixed effects} (assumed to be independent) and \emph{random effects} (assumed to be sampled from a shared distribution).
For example, consider a simple regression model predicting a movie's rating $y$ as a linear combination of features $\mathbf{X}$ (e.g. genre, title): $\hat{y} \sim \mathcal{N}(\beta\mathbf{X}, \epsilon)$ where $\epsilon$ is an error term.
If multiple ratings are provided by each user $j$, they should not be treated as independent--- some users may be more critical and give out lower ratings overall than other users.
It is common to account for this clustered variance by fitting random intercepts and slopes for each user $j$:
$$
    \begin{array}{rl}
        \hat{y}_j\sim &  \mathcal{N}(\beta_j \mathbf{X}_j, \epsilon)\\
        \beta_j \sim  & \mathcal{N}(\mu, \sigma)
    \end{array}
$$
where $\mu$ represents the central tendency shared across the distribution of users, and $\sigma$ represents the population variability. 
This model effectively regularizes user-specific predictions as function of sample size by pulling estimates toward the high density regions of the population distribution.
If a particular user is an outlier, then as more observations are obtained from that user, the more the model will ignore the central tendency and use a user-specific model.
However, if a new user is introduced from the same population, then the central tendency of the random effect provides the best initial guess for their parameters.

% \paragraph{Fixed effects via prefix-tuning.}

% We begin by observing that domain-specific prefixes are equivalent to fixed effects in the mixed-effects framework.
% Specifically, we prepend to the input $x$ a sequence of prefix tokens $\textbf{p}=[p_1, \dots, p_k]$ representing group-specific features (for example, with language taken from movie scripts group-specific features could correspond to a film's genre or title).
% These tokens are then mapped to a corresponding sequence of continuous activations $\textbf{h} = f_\theta(\textbf{p})$ which act as ``virtual tokens'' that control downstream predictions of a language model with frozen parameters.
% We use an MLP for $f_\theta$, which takes the concatenated vector of prefixes as input (\autoref{sec:mlparch} considers independent MLPs for each feature).

\paragraph{Fixed effects via prefix-tuning.}
While mixed-effects models are straightforwardly generalized to non-linear linking functions and non-Gaussian distributions \cite{bates2014fitting,lindstrom1990nonlinear} or cases with multiple nested or cross-cutting groups \cite{baayen2008mixed}, it has been less clear how they could be applied when natural language is the independent variable. 
We begin investigating this problem by considering how to implement a purely fixed-effect language model, where independent group-specific parameters are learned. To represent language data sourced from movie scripts, parameters could be instantiated for each contextual feature to account for clustered variance (e.g. source corpus, genre, and title). 
Each feature would take different values corresponding to different parameters (e.g. ``horror'', ``action'', or ``fantasy'' for genre-level features).

\begin{table*}
\begin{center} 
\caption{Average log perplexity (with 95\% confidence interval) on test set. Our mixed-effects transformers method is able to achieve significantly better performance on contexts it has previously observed (unlike complete-pooling models) while still generalizing well to unseen contexts (unlike no-pooling models).} 
\label{ppl}
\vskip 0.12in
\begin{tabular}{l|cc|cc|cc} 
\hline
\multicolumn{1}{c}{\multirow{2}{*}{\textbf{Model}}} & \multicolumn{2}{|c|}{\textbf{Product Reviews}} & \multicolumn{2}{c|}{\textbf{Reddit Corpus}}  & \multicolumn{2}{c}{\textbf{Movie Corpus}}  \\
& \textbf{Seen} & \textbf{Unseen} & \textbf{Seen} & \textbf{Unseen} & \textbf{Seen} & \textbf{Unseen}\\
\hline
\hline
 Fine-tuning (No Pool) 
  & 3.78$\pm.01$  & 4.23$\pm.01$  
  & 4.03$\pm.01$ & 4.22$\pm.01$
  & 3.83$\pm.05$ & 3.87$\pm.01$ \\
 Fine-tuning (Comp. Pool)
  & 3.72$\pm.01$ & 3.85$\pm.01$ 
  &  4.01$\pm.01$ & 3.93$\pm.01$
  & 3.83$\pm.04$ & 3.87$\pm.01$\\
Conditional Fine-tuning 
  & 3.94$\pm.01$ & 4.19$\pm.03$ 
  & 4.29$\pm.01$ & 4.26$\pm.03$
  & 4.21$\pm.03$ & 4.29$\pm.12$ \\
 Prefix-tuning (No Pool)
  & 3.68$\pm.01$ & 3.97$\pm.01$ 
  & 3.90$\pm.01$ & 3.95$\pm.01$
  & 3.54$\pm.05$ & 3.68$\pm.03$ \\
 Prefix-tuning (Comp. Pool) 
  & 3.79$\pm.01$ & 3.84$\pm.02$ 
   & 4.08$\pm.01$ & 3.83$\pm.03$
  & 3.53$\pm.03$ & 3.65$\pm.11$ \\
Mixed-effects (MET)
  & \bf{3.61}$\pm.01$ & \bf{3.78}$\pm.03$
  & \bf{3.84}$\pm.01$ & \bf{3.80}$\pm.02$
  & \bf{3.47}$\pm.03$ & \bf{3.61}$\pm.12$ \\
\hline
\end{tabular}
\end{center} 
\end{table*}

We generalize the scalar coefficient $\beta_j$ from the regression setting to the language model setting using a set of \emph{prefixes}, $\textbf{p}=[p_1, \dots, p_k]$, which are prepended to the input and yield transformer blocks:
$\textbf{h} = f_\theta(\textbf{p})$ where $\theta$ is a tuneable tensor of parameters.
There are several ways of parameterizing this function; for simplicity, we will take $f_\theta: \mathbb{Z}^k \rightarrow \mathbb{R}^{m \times k}$ to be an embedding layer $W_E$ followed by a series of fully connected layers: 
$$
\textbf{h} = f_\theta(\textbf{p}) = \textrm{MLP}(W_E \cdot \textbf{p}) 
$$
where the dimensionality of the resulting $\textbf{h}$ tensor matches the dimensionality of transformer activations across layers\footnote{For GPT-2, each input token yields an $l \times [k,v]$ tensor, where there are $l = 12$ layers and the dimension of each key and value is 1024.}.
Thus, the prefixes act as ``virtual tokens'' that, like a sequence of input text $x$, control downstream predictions of a language model with frozen parameters $\phi$:
$$
    \hat{y} \sim \textrm{LM}_\phi(x; \textbf{h})
$$ 
Because a single MLP is shared across the full sequence of prefixes, it may be viewed as equivalent to learning interactions between groups in the regression framework (as opposed to a model where each prefix $p_i$ was embedded independently).

% \paragraph{Random effects via regularization.}
% We introduce random effects into the transformer via \emph{hierarchical prefix-tuning}. Critically, instead of assuming that all levels of a feature have independent fixed effects (e.g. that the language associated with one genre is independent of other genres), we assume they are drawn from a common distribution. To accomplish this, we introduce special prefix tokens $\textbf{p}^* = [p^*_1, \dots, p^*_k]$ and randomly replace prefixes with these tokens during training, such that they come to represent the central tendency shared across all levels $j$ of a given feature. Prefixes for specific levels $\textbf{p}^j$ are then regularized toward this shared token:
% $$
% L_\theta(x_j;y) = \log P(y | x_j; \textbf{h}^j)) + \beta || \textbf{h}^j -\textbf{h}^*||^2
% $$
% where $\beta$ is a regularization weight,  $\textbf{h}^j = f_\theta(\textbf{p}^j)$ are the activations derived from the prefixes associated with observation $x_j$.

\paragraph{Random effects via regularization.}
We are now prepared to introduce random effects into the transformer via \emph{hierarchical} prefix-tuning. 
Critically, instead of assuming that all values of a particular feature have independent fixed effects (e.g. that the language associated with one genre is independent of other genres), we would like to assume they are drawn from a common distribution:
$$
\textbf{h} \sim \mathcal{N}(\textbf{h}^*, \beta)
$$ 
where we define $\textbf{h}^*$ to be the activations yielded by a special prefix $\textbf{p}^* = [p^*_0, \dots, p^*_k]$ representing the central tendency across known levels of each feature (see \autoref{pooling}).
In other words, we would like to be able to ``share statistical strength,'' such that our predictions for novel feature values reflect expectations from the entire dataset.

In practice, it is intractable to do probabilistic inference over such a high-dimensional hierarchical neural network, but we may achieve a similar effect via dropout. 
During prefix-tuning, with probability $\epsilon=0.1$, we replace each feature prefix $p_i$ with the corresponding special token $p_i^*$, such that $p_i^*$ comes to reflect the pooled data distribution. This shared token, like $\mu$ in a traditional mixed-effects model, represents the central tendency shared across all values of a particular feature. 
Feature-specific predictions are then regularized toward this shared token by adding a term to the loss function:
$$
L_\theta(x_j;y) = \log P_\phi(y | x_j; f_\theta(\textbf{p}^j)) + \beta || \textbf{h}^j -\textbf{h}^*||^2
$$
where the regularization parameter, $\beta=0.01$ is comparable to the standard deviation for random effects in a typical regression model.

\begin{figure*}[t]
\begin{center}
\includegraphics[width=16cm]{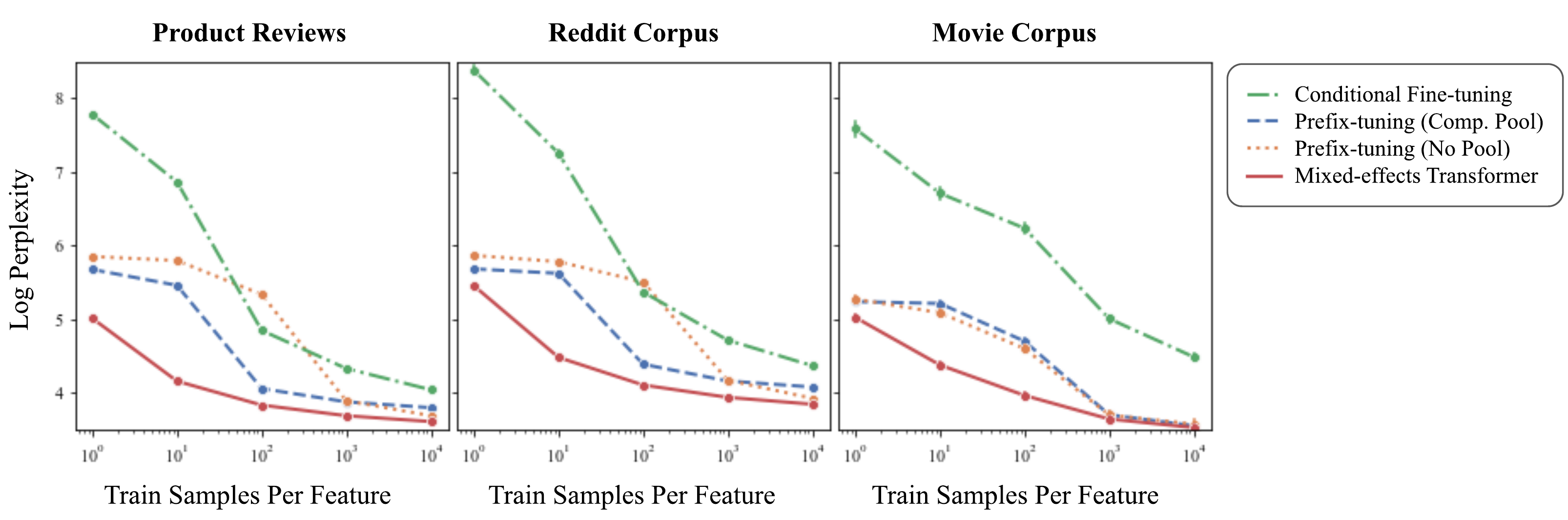}
\end{center}
\caption{Log perplexity (with 95\% confidence interval) on test set after training on different lengths of data for Product Reviews (left), Reddit Corpus (middle), and Movie Corpus (right).}
\label{dataeff}
\end{figure*}

\section{Datasets}
We examine language use across contexts in three distinct domains: product reviews, online posts, and movie dialogue. 100,000 sentences were sampled for training from 10 distinct product categories within the Amazon Customer Reviews Dataset\footnote{https://s3.amazonaws.com/amazon-reviews-pds/readme.html}, a.k.a \textbf{Product Reviews}; 100,000 sentences were sampled from 10 subreddits (subsidiary forums representing distinct topical communities) within the \textbf{Reddit Corpus} \cite{Henderson2019}; and, 10,000 sentences were sampled from 10 genres within the Cornell Movie-Dialogs Corpus \cite{Danescu2011Chameleons}, a.k.a \textbf{Movie Corpus}.
Further information about these datasets and their contextual features can be seen in \autoref{sec:data}.

\section{Results}
We evaluate the ability of the MET to capture language use within known and novel contexts. Further, we assess the data efficiency of our method and its ability to represent complex contexts with multiple relevant features. 
We compare the performance of our approach against several baselines.
In the complete-pooling and no-pooling variants of prefix-tuning we ablate different components, only learning a single prefix shared across all features, or only learning independent prefixes, respectively.
We also compare a traditional domain adaptation approach, where we omit prefixes and fine-tune the transformer end-to-end either on the entire dataset (complete pooling) or for each feature separately (no pooling).
Finally, we compare our method against \emph{conditional fine-tuning}, where a string representing the prefix text (e.g. \texttt{[corpus] movie\_dialogue [genre] horror} ) is prepended to the input and the model is fine-tuned end-to-end. See \autoref{sec:expt} for additional details.

\subsection{Adaptation to known contexts}

We begin by evaluating MET on a standard cross-domain language modeling task.
Examples from each contextual feature (e.g. genres) are seen during training and we assess the model's predictions on held-out examples from those contexts.  
This task evaluates the extent to which explicitly modeling multiple sources of extra-textual variance may improve a model's ability to predict further language across those diverse sources.
\autoref{ppl} (left column) shows the log perplexity of each method.
First, replicating \citet{Li2021}, we find that prefix-tuning generally outperforms end-to-end fine-tuning.
Second, as expected, pure \emph{no pooling} models generally out-perform pure \emph{complete pooling} models; the former is able to learn independent models for each sub-domain while the latter is constrained to learn a single model for the entire corpus.
Third, the conditional fine-tunining method performs particularly poorly, likely due to data sparsity with respect to feature values. 
Finally, METs outperform even the no-pooling baselines on all three datasets, suggesting that replacing fixed effects with random effects enables better adaptation to known domains.
In other words, while massive language models may have difficulty tuning to individual contexts with few samples using traditional methods, mixed-effect prefix-tuning enables them to overcome this limitation by leveraging information gained about language use in other contexts.

\subsection{Generalization to novel contexts}

Next, we evaluate our method's ability to generalize to novel, \emph{unseen} contexts, where traditional domain adaptation methods typically do poorly.
We evaluate on a test set containing examples with contextual feature values that were entirely held-out of the training set (\autoref{ppl}, right column).
We find that the complete-pooling models typically generalize better to new features than no-pooling models; the former have seen more data across a broader spectrum of feature values during training, whereas conditional fine-tuning is least successful. METs, which represent unseen feature values with the shared prefix token, attain the best perplexity on all three datasets, capturing feature-specific language without sacrificing the ability to generalize. This performance is likely in part due to the method's ability to discount individual ``outlier'' features from affecting the overall distribution, a key aspect of Bayesian hierarchical modelling. It is worth noting that models occasionally achieve better performance on unseen features likely due to a quirk of the split: the predictability of language can vary significantly across feature values.

\subsection{Data efficiency}

A well-known benefit of mixed-effects models in classical regression settings is their ability to flexibly interpolate as a function of sample size.
As more observations become available, they allow domain-specific predictions to deviate more strongly from the central tendency of the population. 
To better evaluate performance as a function of sample size, we construct training sets of different sizes, interpolating between settings where the model has only seen one example of a given feature up to cases where it sees many thousands of examples  (\autoref{dataeff}).
In lower-data settings, the complete pooling approaches outperform no pooling approaches, as the no-pooling model is making predictions based on only a handful of examples. 
As the amount of data per feature increases, no-pooling method eventually achieve better performance. 
Meanwhile, the MET consistently outperforms both pooling methods.
Particularly in low-data settings, this approach is able to make feature-specific inferences without sacrificing knowledge acquired from other features.

\subsection{Adaptation to multi-feature contexts} 
Finally, one of the most intriguing properties of mixed-effects models is their ability to account for not just a single ``domain'' feature but multiple cross-cutting features in different combinations. 
We assess the ability of METs to represent language in  complex contexts where multiple contextual features are available. 
More significant performance improvements are realized in less sparse feature spaces, so we run this evaluation on a subset of the data with dense secondary contextual features (product, user, and movie) which are taken from the top 10 values occurring within each of the top 10 primary features (product category, subreddit, and movie genre).
In \autoref{context-comp} we compare the change in log perplexity when observing only one contextual feature to observing a secondary feature and find that including multiple feature prefixes improves performance.

\begin{table}[h]
\begin{center} 
\caption{Log perplexity while observing only one contextual feature versus multiple contextual features.
} 
\label{context-comp}
\vskip 0.12in
\begin{tabular}{l|cc} 
\hline
\textbf{Dataset} & \textbf{Single-feature} & \textbf{Multi-feature} \\
\hline
\hline
Amazon & 3.47$\pm.03$ & 3.33$\pm.03$\\
Reddit & 3.40$\pm.04$ & 3.26$\pm.05$ \\
Movies & 3.29$\pm.03$ & 3.07$\pm.04$\\
\hline
\end{tabular}
\end{center} 
\end{table}

\subsection{Comparison to fine-tuned adapters}
In recent work, context-specific adapters--- lightweight layers added after each transformer block--- have been successfully utilized for hierarchical adaptation. In \citet{chronopoulou2022efficient} internet domains from Common Crawl's colossal, cleaned web crawl corpus, \textbf{C4} \cite{Henderson2019}, are modelled as a tree structure with individual adapters associated to each node.
In \autoref{adapter}, we compare this method with our approach after training on 100,000 sentences from 10 web domains\footnote{fronteirsin.org, chicagotribune.com, link.springer.com, aljazeera.com, instructables.com, npr.org, dailymail.co.uk, csmonitor.com, baltimoresun.com, city-data.com} each. While both models demonstrate similar performance boosts for in-distribution language data, the MET sees improved performance modelling out-of-distribution language--- offering an effective alternative solution to hierarchical adaptation in low resource settings.

\begin{table}[h]
\begin{center} 
\caption{Average log perplexity on the C4 test set.} 
\label{adapter}
\vskip 0.12in
\begin{tabular}{l|cc} 
\hline
\textbf{Model} & \textbf{Seen} & \textbf{Unseen} \\
\hline
\hline
Fine-tuning (Comp. Pool) & 3.89 & 4.00 \\
Mixed-effects (MET) & 3.76 & 3.92 \\
Hierarchical Adapters & 3.76 & 4.34 \\
\hline
\end{tabular}
\end{center} 
\end{table}

\section{Conclusion}

Human language is flexible, and people are able to adapt their expectations to many aspects of context, from speaker identity to the conversational setting. In this paper, we introduce mixed-effects transformers (METs) as an effective method of adapting to hierarchically structured domains of language use across labeled contextual features.
Beyond language modeling, this approach may be useful for controlled generation and more qualitative analyses of what makes certain features distinctive (see \autoref{sec:prefixchar} for preliminary analysis). 

\section{Limitations}

We were not able to investigate how our method scales to larger feature sets (e.g. the tens of thousands of product IDs in Product Reviews), due to constraints on compute (we use an NVIDIA TITAN X GPU for all experiments).
We expect there is a point where the parameter budget of the prefixes and MLP grows larger than the frozen model, which would require alternative parameterizations. % For example, not computing unique prefix values for every layer of the transformer stack.
Additionally, our regularization technique only affects prefixes within batches, so batch size and composition may affect the learning of $p^*$ central tendencies.  

\section{Acknowledgements}
This research was supported in part by the Stanford HAI Hoffman--Yee project `Towards grounded, adaptive communication agents'. 
RDH was funded by a C.V. Starr Postdoctoral Fellowship and NSF SPRF award $\#$1911835. We are also grateful for helpful conversations and feedback from members of the Computation $\&$ Cognition Lab, the Princeton NLP Group, and our review committee.

\bibliography{custom}

\begin{thebibliography}{39}
\expandafter\ifx\csname natexlab\endcsname\relax\def\natexlab#1{#1}\fi

\bibitem[{Baayen et~al.(2008)Baayen, Davidson, and Bates}]{baayen2008mixed}
R~Harald Baayen, Douglas~J Davidson, and Douglas~M Bates. 2008.
\newblock Mixed-effects modeling with crossed random effects for subjects and
  items.
\newblock \emph{Journal of memory and language}, 59(4):390--412.

\bibitem[{Baltagi(2008)}]{Baltagi2021}
Badi~H Baltagi. 2008.
\newblock \emph{Econometric analysis of panel data}, volume~4.
\newblock Springer.

\bibitem[{Bates et~al.(2014)Bates, M{\"a}chler, Bolker, and
  Walker}]{bates2014fitting}
Douglas Bates, Martin M{\"a}chler, Ben Bolker, and Steve Walker. 2014.
\newblock Fitting linear mixed-effects models using lme4.
\newblock \emph{arXiv preprint arXiv:1406.5823}.

\bibitem[{Blodgett et~al.(2016)Blodgett, Green, and
  O'Connor}]{blodgett2016demographic}
Su~Lin Blodgett, Lisa Green, and Brendan O'Connor. 2016.
\newblock Demographic dialectal variation in social media: A case study of
  african-american english.
\newblock \emph{arXiv preprint arXiv:1608.08868}.

\bibitem[{Bommasani et~al.(2021)Bommasani, Hudson, Adeli, Altman, Arora, von
  Arx, Bernstein, Bohg, Bosselut, Brunskill
  et~al.}]{bommasani2021opportunities}
Rishi Bommasani, Drew~A Hudson, Ehsan Adeli, Russ Altman, Simran Arora, Sydney
  von Arx, Michael~S Bernstein, Jeannette Bohg, Antoine Bosselut, Emma
  Brunskill, et~al. 2021.
\newblock On the opportunities and risks of foundation models.
\newblock \emph{arXiv preprint arXiv:2108.07258}.

\bibitem[{Brown et~al.(2020)Brown, Mann, Ryder, Subbiah, Kaplan, Dhariwal,
  Neelakantan, Shyam, Sastry, Askell, Agarwal, Herbert-Voss, Krueger, Henighan,
  Child, Ramesh, Ziegler, Wu, Winter, Hesse, Chen, Sigler, Litwin, Gray, Chess,
  Clark, Berner, McCandlish, Radford, Sutskever, and Amodei}]{Brown2020}
Tom~B. Brown, Benjamin Mann, Nick Ryder, Melanie Subbiah, Jared Kaplan,
  Prafulla Dhariwal, Arvind Neelakantan, Pranav Shyam, Girish Sastry, Amanda
  Askell, Sandhini Agarwal, Ariel Herbert-Voss, Gretchen Krueger, Tom Henighan,
  Rewon Child, Aditya Ramesh, Daniel~M. Ziegler, Jeffrey Wu, Clemens Winter,
  Christopher Hesse, Mark Chen, Eric Sigler, Mateusz Litwin, Scott Gray,
  Benjamin Chess, Jack Clark, Christopher Berner, Sam McCandlish, Alec Radford,
  Ilya Sutskever, and Dario Amodei. 2020.
\newblock Language models are few-shot learners.

\bibitem[{Chronopoulou et~al.(2022)Chronopoulou, Peters, and
  Dodge}]{chronopoulou2022efficient}
Alexandra Chronopoulou, Matthew Peters, and Jesse Dodge. 2022.
\newblock \href {https://doi.org/10.18653/v1/2022.naacl-main.96} {Efficient
  hierarchical domain adaptation for pretrained language models}.
\newblock In \emph{Proceedings of the 2022 Conference of the North American
  Chapter of the Association for Computational Linguistics: Human Language
  Technologies}, pages 1336--1351, Seattle, United States. Association for
  Computational Linguistics.

\bibitem[{Clark(1998)}]{Clark1998-CLACL}
Herbert~H. Clark. 1998.
\newblock Communal lexicons.
\newblock In Kirsten Malmkjaer and John Williams, editors, \emph{Context in
  Language Learning and Language Understanding}, page~63. Cambridge University
  Press.

\bibitem[{Dai et~al.(2020)Dai, Karimi, Hachey, and Paris}]{dai-etal-2020-cost}
Xiang Dai, Sarvnaz Karimi, Ben Hachey, and Cecile Paris. 2020.
\newblock \href {https://doi.org/10.18653/v1/2020.findings-emnlp.151}
  {Cost-effective selection of pretraining data: A case study of pretraining
  {BERT} on social media}.
\newblock In \emph{Findings of the Association for Computational Linguistics:
  EMNLP 2020}, pages 1675--1681, Online. Association for Computational
  Linguistics.

\bibitem[{Danescu-Niculescu-Mizil and Lee(2011)}]{Danescu2011Chameleons}
Cristian Danescu-Niculescu-Mizil and Lillian Lee. 2011.
\newblock Chameleons in imagined conversations: A new approach to understanding
  coordination of linguistic style in dialogs.
\newblock In \emph{Proceedings of the Workshop on Cognitive Modeling and
  Computational Linguistics, ACL 2011}.

\bibitem[{Daume~III and Marcu(2006)}]{daume2006domain}
Hal Daume~III and Daniel Marcu. 2006.
\newblock Domain adaptation for statistical classifiers.
\newblock \emph{Journal of artificial Intelligence research}, 26:101--126.

\bibitem[{Eckert(2012)}]{eckert2012three}
Penelope Eckert. 2012.
\newblock Three waves of variation study: The emergence of meaning in the study
  of sociolinguistic variation.
\newblock \emph{Annual review of Anthropology}, 41:87--100.

\bibitem[{Feldman et~al.(2021)Feldman, Tiwari, Cheah, Foulds, and
  Pan}]{feldman2021analyzing}
Philip Feldman, Sim Tiwari, Charissa~SL Cheah, James~R Foulds, and Shimei Pan.
  2021.
\newblock Analyzing covid-19 tweets with transformer-based language models.
\newblock \emph{arXiv preprint arXiv:2104.10259}.

\bibitem[{Flek(2020)}]{flek2020returning}
Lucie Flek. 2020.
\newblock Returning the n to nlp: Towards contextually personalized
  classification models.
\newblock In \emph{Proceedings of the 58th Annual Meeting of the Association
  for Computational Linguistics}, pages 7828--7838.

\bibitem[{Gelman and Hill(2006)}]{gelman2006data}
Andrew Gelman and Jennifer Hill. 2006.
\newblock \emph{Data analysis using regression and multilevel/hierarchical
  models}.
\newblock Cambridge university press.

\bibitem[{Hambardzumyan et~al.(2021)Hambardzumyan, Khachatrian, and
  May}]{hambardzumyan2021warp}
Karen Hambardzumyan, Hrant Khachatrian, and Jonathan May. 2021.
\newblock Warp: Word-level adversarial reprogramming.
\newblock \emph{arXiv preprint arXiv:2101.00121}.

\bibitem[{Hawkins et~al.(2022)Hawkins, Franke, Frank, Goldberg, Smith,
  Griffiths, and Goodman}]{hawkins2021partners}
Robert Hawkins, Michael Franke, Michael~C Frank, Adele Goldberg, Kenny Smith,
  Thomas~L Griffiths, and Noah~D Goodman. 2022.
\newblock From partners to populations: A hierarchical bayesian account of
  coordination and convention.
\newblock \emph{Psychological Review}.

\bibitem[{Henderson et~al.(2019)Henderson, Budzianowski, Casanueva, Coope,
  Gerz, Kumar, Mrk{\v{s}}i\'c, Spithourakis, Su, Vulic, and
  Wen}]{Henderson2019}
Matthew Henderson, Pawe{\l} Budzianowski, I{\~{n}}igo Casanueva, Sam Coope,
  Daniela Gerz, Girish Kumar, Nikola Mrk{\v{s}}i\'c, Georgios Spithourakis,
  Pei-Hao Su, Ivan Vulic, and Tsung-Hsien Wen. 2019.
\newblock \href {https://arxiv.org/abs/1904.06472} {A repository of
  conversational datasets}.
\newblock In \emph{Proceedings of the Workshop on {NLP} for Conversational
  {AI}}.
\newblock Data available at github.com/PolyAI-LDN/conversational-datasets.

\bibitem[{Hofmann et~al.(2022)Hofmann, Glava{\v{s}}, Ljube{\v{s}}i{\'c},
  Pierrehumbert, and Sch{\"u}tze}]{hofmann2022geographic}
Valentin Hofmann, Goran Glava{\v{s}}, Nikola Ljube{\v{s}}i{\'c}, Janet~B
  Pierrehumbert, and Hinrich Sch{\"u}tze. 2022.
\newblock Geographic adaptation of pretrained language models.
\newblock \emph{arXiv preprint arXiv:2203.08565}.

\bibitem[{Hovy(2015)}]{hovy2015demographic}
Dirk Hovy. 2015.
\newblock Demographic factors improve classification performance.
\newblock In \emph{Proceedings of the 53rd annual meeting of the Association
  for Computational Linguistics and the 7th international joint conference on
  natural language processing (volume 1: Long papers)}, pages 752--762.

\bibitem[{Lazaridou et~al.(2021)Lazaridou, Kuncoro, Gribovskaya, Agrawal,
  Liska, Terzi, Gimenez, de~Masson~d'Autume, Kocisky, Ruder
  et~al.}]{lazaridou2021mind}
Angeliki Lazaridou, Adhi Kuncoro, Elena Gribovskaya, Devang Agrawal, Adam
  Liska, Tayfun Terzi, Mai Gimenez, Cyprien de~Masson~d'Autume, Tomas Kocisky,
  Sebastian Ruder, et~al. 2021.
\newblock Mind the gap: Assessing temporal generalization in neural language
  models.
\newblock \emph{Advances in Neural Information Processing Systems}, 34.

\bibitem[{Lee et~al.(2020)Lee, Yoon, Kim, Kim, Kim, So, and
  Kang}]{lee2020biobert}
Jinhyuk Lee, Wonjin Yoon, Sungdong Kim, Donghyeon Kim, Sunkyu Kim, Chan~Ho So,
  and Jaewoo Kang. 2020.
\newblock Biobert: a pre-trained biomedical language representation model for
  biomedical text mining.
\newblock \emph{Bioinformatics}, 36(4):1234--1240.

\bibitem[{Lester et~al.(2021)Lester, Al-Rfou, and Constant}]{lester2021power}
Brian Lester, Rami Al-Rfou, and Noah Constant. 2021.
\newblock The power of scale for parameter-efficient prompt tuning.
\newblock In \emph{Proceedings of the 2021 Conference on Empirical Methods in
  Natural Language Processing}, pages 3045--3059.

\bibitem[{Li and Liang(2021)}]{Li2021}
Xiang~Lisa Li and Percy Liang. 2021.
\newblock Prefix-tuning: Optimizing continuous prompts for generation.
\newblock In \emph{Proceedings of the 59th Annual Meeting of the Association
  for Computational Linguistics and the 11th International Joint Conference on
  Natural Language Processing (Volume 1: Long Papers)}, pages 4582--4597.

\bibitem[{Lindstrom and Bates(1990)}]{lindstrom1990nonlinear}
Mary~J Lindstrom and Douglas~M Bates. 1990.
\newblock Nonlinear mixed effects models for repeated measures data.
\newblock \emph{Biometrics}, pages 673--687.

\bibitem[{Linzen and Baroni(2021)}]{linzen2021syntactic}
Tal Linzen and Marco Baroni. 2021.
\newblock Syntactic structure from deep learning.
\newblock \emph{Annual Review of Linguistics}, 7:195--212.

\bibitem[{Luu et~al.(2021)Luu, Khashabi, Gururangan, Mandyam, and
  Smith}]{luu2021time}
Kelvin Luu, Daniel Khashabi, Suchin Gururangan, Karishma Mandyam, and Noah~A
  Smith. 2021.
\newblock Time waits for no one! analysis and challenges of temporal
  misalignment.
\newblock \emph{arXiv preprint arXiv:2111.07408}.

\bibitem[{Nardy et~al.(2013)Nardy, Chevrot, and Barbu}]{nardy2013acquisition}
Aur{\'e}lie Nardy, Jean-Pierre Chevrot, and St{\'e}phanie Barbu. 2013.
\newblock The acquisition of sociolinguistic variation: Looking back and
  thinking ahead.
\newblock \emph{Linguistics}, 51(2):255--284.

\bibitem[{Nguyen et~al.(2020)Nguyen, Vu, and Nguyen}]{nguyen2020bertweet}
Dat~Quoc Nguyen, Thanh Vu, and Anh~Tuan Nguyen. 2020.
\newblock Bertweet: A pre-trained language model for english tweets.
\newblock \emph{arXiv preprint arXiv:2005.10200}.

\bibitem[{Ostapenko et~al.(2022)Ostapenko, Wintner, Fricke, and
  Tsvetkov}]{Ostapenko2022}
Alissa Ostapenko, Shuly Wintner, Melinda Fricke, and Yulia Tsvetkov. 2022.
\newblock Speaker information can guide models to better inductive biases: A
  case study on predicting code-switching.
\newblock In \emph{Proceedings of the 60th Annual Meeting of the Association
  for Computational Linguistics}, pages 4582--4597.

\bibitem[{R{\"o}ttger and Pierrehumbert(2021)}]{rottger2021temporal}
Paul R{\"o}ttger and Janet~B Pierrehumbert. 2021.
\newblock Temporal adaptation of bert and performance on downstream document
  classification: Insights from social media.
\newblock \emph{arXiv preprint arXiv:2104.08116}.

\bibitem[{Sawhney et~al.(2020)Sawhney, Joshi, Gandhi, and
  Shah}]{sawhney2020time}
Ramit Sawhney, Harshit Joshi, Saumya Gandhi, and Rajiv Shah. 2020.
\newblock A time-aware transformer based model for suicide ideation detection
  on social media.
\newblock In \emph{Proceedings of the 2020 conference on empirical methods in
  natural language processing (EMNLP)}, pages 7685--7697.

\bibitem[{Schlechtweg et~al.(2019)Schlechtweg, H{\"a}tty, Del~Tredici, and
  Schulte~im Walde}]{schlechtweg-etal-2019-wind}
Dominik Schlechtweg, Anna H{\"a}tty, Marco Del~Tredici, and Sabine Schulte~im
  Walde. 2019.
\newblock \href {https://doi.org/10.18653/v1/P19-1072} {A wind of change:
  Detecting and evaluating lexical semantic change across times and domains}.
\newblock In \emph{Proceedings of the 57th Annual Meeting of the Association
  for Computational Linguistics}, pages 732--746, Florence, Italy. Association
  for Computational Linguistics.

\bibitem[{Wang et~al.(2019)Wang, Pruksachatkun, Nangia, Singh, Michael, Hill,
  Levy, and Bowman}]{wang2019superglue}
Alex Wang, Yada Pruksachatkun, Nikita Nangia, Amanpreet Singh, Julian Michael,
  Felix Hill, Omer Levy, and Samuel Bowman. 2019.
\newblock Superglue: A stickier benchmark for general-purpose language
  understanding systems.
\newblock \emph{Advances in neural information processing systems}, 32.

\bibitem[{Warstadt et~al.(2019)Warstadt, Singh, and
  Bowman}]{warstadt2019neural}
Alex Warstadt, Amanpreet Singh, and Samuel~R Bowman. 2019.
\newblock Neural network acceptability judgments.
\newblock \emph{Transactions of the Association for Computational Linguistics},
  7:625--641.

\bibitem[{Wolf et~al.(2019)Wolf, Debut, Sanh, Chaumond, Delangue, Moi, Cistac,
  Rault, Louf, Funtowicz, and Brew}]{huggingface}
Thomas Wolf, Lysandre Debut, Victor Sanh, Julien Chaumond, Clement Delangue,
  Anthony Moi, Pierric Cistac, Tim Rault, R{\'{e}}mi Louf, Morgan Funtowicz,
  and Jamie Brew. 2019.
\newblock \href {http://arxiv.org/abs/1910.03771} {Huggingface's transformers:
  State-of-the-art natural language processing}.
\newblock \emph{CoRR}, abs/1910.03771.

\bibitem[{Yang and Eisenstein(2017)}]{yang2017overcoming}
Yi~Yang and Jacob Eisenstein. 2017.
\newblock Overcoming language variation in sentiment analysis with social
  attention.
\newblock \emph{Transactions of the Association for Computational Linguistics},
  5:295--307.

\bibitem[{Zeng et~al.(2020)Zeng, Yang, Ju, Yang, Wang, Zhang, Zhou, Zeng, Dong,
  Zhang, Fang, Zhu, Chen, and Xie}]{zeng-etal-2020-meddialog}
Guangtao Zeng, Wenmian Yang, Zeqian Ju, Yue Yang, Sicheng Wang, Ruisi Zhang,
  Meng Zhou, Jiaqi Zeng, Xiangyu Dong, Ruoyu Zhang, Hongchao Fang, Penghui Zhu,
  Shu Chen, and Pengtao Xie. 2020.
\newblock \href {https://doi.org/10.18653/v1/2020.emnlp-main.743}
  {{M}ed{D}ialog: Large-scale medical dialogue datasets}.
\newblock In \emph{Proceedings of the 2020 Conference on Empirical Methods in
  Natural Language Processing (EMNLP)}, pages 9241--9250, Online. Association
  for Computational Linguistics.

\bibitem[{Zhou et~al.(2021)Zhou, Yang, Loy, and Liu}]{zhou2021learning}
Kaiyang Zhou, Jingkang Yang, Chen~Change Loy, and Ziwei Liu. 2021.
\newblock Learning to prompt for vision-language models.
\newblock \emph{arXiv preprint arXiv:2109.01134}.

\end{thebibliography}
\bibliographystyle{acl_natbib}

\appendix
\clearpage

\section{Datasets}
\label{sec:data}

We assess the performance of the MET on three datasets: the Amazon Customer Reviews Dataset, Reddit Corpus, and the Cornell Movie-Dialogs Corpus.

The Amazon Customer Reviews Dataset (\textbf{Product Reviews}) compiles reviews across product categories. We sampled 100,000 sentences from reviews in 11 product categories: video games, pet products, grocery, home, electronics, beauty, baby, automotive, apparel, books, and sports (which was held-out during training). In addition to product category, the metadata for Product Reviews also includes a product id. 

The \textbf{Reddit Corpus} is a collection of posts and comments from different subreddits (subsidiary forums representing distinct topical communities) from the popular social media site Reddit. We sampled 100,000 sentences from posts and comments in 11 subreddits: aww, todayilearned, apple, pokemontrades, relationship\_advice, DebateReligion, worldnews, nba, Naruto, hiphopheads, and AskReddit (which was held-out during training). The metadata for Reddit posts also the username of the poster.

The Cornell Movie-Dialogs Corpus (\textbf{Movie Corpus}) is a dataset of
movie dialogue for a number of genres. We sampled 10,000 sentences\footnote{The number of sentences available within each individual genre does not exceed 100,000 in Movie Corpus.} of dialogue from 11 genres: action, adventure, comedy, crime, drama, horror, mystery, romance, sci-fi, thriller, and fantasy (which was held-out during training). The metadata for this dataset also includes the movie title. 

We used a 80/10/10 train-val-test split in addition to the test sentences sampled from the aforementioned held-out feature values (e.g., movie dialogue from the fantasy genre) which were used in the evaluation of our models for unseen contexts.

\begin{figure*}[t!]
\begin{center}
\includegraphics[width=16cm]{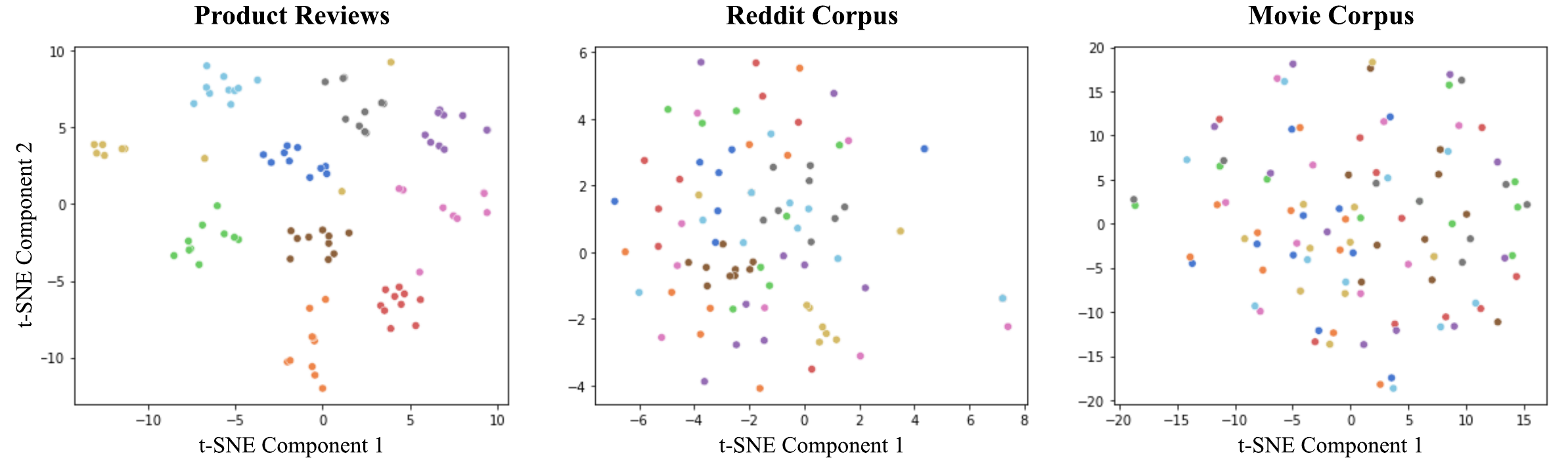}
\end{center}
\caption{t-SNE dimensionality reduction of secondary contextual feature (e.g. movie title) prefix embeddings color coded according to primary feature (e.g. genre).}
\label{tsne}
\end{figure*}

\section{Experimental setup}
\label{sec:expt}

We assigned each individual contextual feature value a unique prefix token, which could take on 128 values. In all experiments, the first prefix represents the overall corpus or task (e.g., Movie Corpus), and the following prefixes represent successively more fine-grained contextual features (e.g. genre and movie title).

The MLPs used to recover prefixes from feature values consisted of 2 layers with a hidden dimension of 512 and took input from an embedding layer with an embedding size of 512. 
The dimensionality of the MLP's output tensor matches the dimensionality of the language model's transformer activations across layers. For the language model we use GPT-2, where each input token yields an $l \times [k,v]$ tensor with $l = 12$ layers and the dimension of each key and value is 1024. 

Our implementations are based on the Hugging Face Transformer models \cite{huggingface}. Our models were trained with a learning rate of $0.00001$ using the AdamW optimizer and a batch size of 4 when sampling utterances.

\section{Shared vs. independent prefix MLP}
\label{sec:mlparch}

\begin{table}[H]
\begin{center} 
\caption{Log perplexity on Movie Corpus for shared prefix MLP and independent prefix MLP architectures on test set.} 
\label{arch-comp}
\vskip 0.12in
\begin{tabular}{lc} 
\hline
\multicolumn{1}{c}{\textbf{Architecture}} & \textbf{Log Perplexity} \\
\hline
\hline
Shared Prefix MLP & 3.61 (3.61, 3.62) \\
Independent Prefix MLP & 3.61 (3.60, 3.62) \\
\hline
\end{tabular}
\end{center} 
\end{table}

We tested two hierarchical prefix architectures on Product Reviews for models containing two prefixes: a corpus-level prefix and a product-category-level prefix. The first, the shared prefix MLP architecture, uses one MLP to produce all feature prefixes and thereby allows information to be shared across features. The second, the independent prefix MLP architecture, uses multiple independent MLPs to produce a prefix for each feature. Assessment of the log perplexity of both methods reveals negligible difference in performance (see \autoref{arch-comp}). Ultimately, the shared prefix MLP architecture was chosen for our MET approach as this method requires less resources during training.

\section{Characterization of the prefix space}
\label{sec:prefixchar}

\subsection{Distinctive utterances sampled from feature prefixes}

\begin{table}[t]
\begin{center} 
\caption{Utterances from Product Reviews test data with the highest difference in perplexity when the model's prefix corresponds to the given Amazon product category.} 
\label{select}
\vskip 0.12in
\begin{tabular}{ll} 
\hline
\textbf{Product Category} & \textbf{Sentence} \\
\hline
\hline
Apparel & Great shirt \\
Automotive & Good fit \\
Baby & Great crib \\
Beauty & Great scent \\
Books & good autobiography \\
Electronics & good sound \\
Grocery & Excellent coffee \\
Home & Love this vacuum!! \\
Pet Products & fun toy \\
Video Games & great game \\
\hline
\end{tabular}
\end{center} 
\end{table}

\begin{table}[h!]
\begin{center} 
\caption{Utterances generated from the prompt "I love" using subreddit-specific prefixes learned on Reddit Corpus.} 
\label{gen}
\vskip 0.12in
\begin{tabular}{ll} 
\hline
\textbf{Subreddit} & \textbf{Sentence} \\
\hline
\hline
% apple & I love the fact that you can \\
% & use the same app to access \\
% & your phone's camera. \\
apple & I love the iPhone \\
aww & I love the way he looks. \\
naruto & I love Izumi \\
% pokemontrades & I love the idea of a \\
% & "diamond" in the middle of \\
% & a game. \\
nba & I love the way he's playing. \\
\hline
\end{tabular}
\end{center} 
\end{table}

To better understand the specific linguistic differences that our model uses to make better predictions, we queried the model for distinctive sentences.
Specifically, we searched the training data for sentences with the highest difference in perplexity for a given feature compared to other features. 
We expected distinctive utterances to contain language that is common for the given feature value while being uncommon for other feature values.
In \autoref{select}, we show the most distinctive utterances found to correspond to the different product category prefixes for Product Reviews. 
We see that the prefixes have successfully learned to represent distinctive language used in each domain (e.g. ``shirt'' for apparel and ``autobiography'' for books).
In this case, the product category features are already easily interpretable, so these utterances may be unsurprising. 
However, we believe that this method may enable interpretation of less legible features in other datasets (e.g. identifying different subcommunities in social networks by clustering prefixes.)

\subsection{Prompted generations from feature prefixes}

To directly observe the linguistic trends our model picks up on within specific contexts, we prompted our model generate utterances corresponding to specific feature values. 
We expect generated utterances to contain language typical of the domains invoked in prefix selection.
In \autoref{gen}, we show generated utterances for a handful of subreddit prefixes trained on Reddit Corpus.
We find that these prefixes contain enough contextual signal to cater the generated utterances to their respective domains (e.g. the mention of ``iPhone'' within the apple subreddit generation).

\subsection{t-SNE analysis of feature prefixes}

We perform a dimensionality reduction on the secondary contextual feature (movie title, username, product id) prefix embeddings to reveal the learned structure of our datasets. Specifically, we use t-distributed stochastic neighbor embedding (t-SNE) to map the high-dimensionality prefix embeddings to a location in a two-dimensional map. After color coding the resulting two-dimensional points according to their primary feature (genre, subreddit, product category), we observe that prefix embeddings cluster differently in accordance with each dataset's underlying structure (see \autoref{tsne}). Reddit and Movie Corpus do not have strongly correlated clusters of features because the underlying structure of the data is cross-cut with respect to the features represented: users frequently post in multiple subreddits and movie titles often simultaneously belong to many genres. This behavior is expected as a mixed-effects model should effectively partition off correlations between cross-cut features.
On the other had, when features are perfectly nested, as in Product Reviews where a specific product belongs to only one product category, we see an expected clustering of product prefixes according to their category.

\end{document}